# A Constraint-Handling Technique for Genetic Algorithms using a Violation Factor

[1,2]Adam Chehouri, [2]Rafic Younes, [1]Jean Perron and [3]Adrian Ilinca

[1]Anti Icing Materials International Laboratory (AMIL),
Université du Québec à Chicoutimi, 555 boulevard de l'Université, Canada G7H 2B1, Canada
[2]Faculty of Engineering, Third Branch, Lebanese University, Rafic Harriri Campus, Hadath, Beirut, Lebanon
[3]Wind Energy Research Laboratory (WERL),
Université du Québec à Rimouski, 300 allée des Ursulines, Québec, Canada G5L 3A1, Canada

**Abstract:** Over the years, several meta-heuristic algorithms were proposed and are now emerging as common methods for constrained optimization problems. Among them, genetic algorithms (GA's) shine as popular evolutionary algorithms (EA's) in engineering optimization. Most engineering design problems are difficult to resolve with conventional optimization algorithms because they are highly nonlinear and contain constraints. In order to handle these constraints, the most common technique is to apply penalty functions. The major drawback is that they require tuning of parameters, which can be very challenging. In this paper, we present a constraint-handling technique for GA's solely using the violation factor, called VCH (Violation Constraint-Handling) method. Several benchmark problems from the literature are examined. The VCH technique was able to provide a consistent performance and match results from other GA-based techniques.

## Introduction

Many optimization problems in engineering are highly nonlinear, containing a mixture of discrete and continuous design variables subject to a series of constraints. Such problems are known as constrained optimization problems or nonlinear programming problems in which traditional calculus-based methods struggle to solve. These numerical optimization methods are highly deterministic and are convenient in finding the global optimum for simple problems by improving the solution in the vicinity of a starting point. However, they have major drawbacks with complex engineering problems i.e.: difficulty in computing the derivatives, sensitivity to the initial conditions and a large memory requirement.

Because of these downsides, over the years, several heuristic and meta-heuristic algorithms were proposed. They are now emerging as popular methods for the solution of complex engineering problems. These algorithms are purely stochastic and consist of approximate methods but on the contrary are derivative-free techniques. Heuristic methods try to find decent solutions that are easily reachable but are not necessarily the best solutions by means of trial and error. Further developments of heuristics are the so-called meta-heuristic algorithms: A higher level of optimization compared to heuristic algorithms. The meta-heuristic techniques include: genetic algorithms (GA, Holland (1975)), simulated annealing (SA, Kirkpatrick and Vecchi (1983)), particle swarm optimization (PSO, Eberhart and Kennedy (1995)), ant colony optimization (ACO, Dorigo *et al.* (1996)), tabu search (Glover 1977) etc. Among all meta-heuristics, genetic algorithms (proposed by Holland (1975)) are one of the most popular evolutionary algorithms (EA's). By mimicking the basic Darwinian mechanism from the famous book "The Origin of Species" (Darwin and Bynum, 2009) defined *natural selection* of biological systems or the principle of the *survival of the fittest*. GA's try to evolve the population of *chromosomes* that are fitter by applying three key evolutionary operators: selection, crossover and mutation. The attempt is to produce a new generation or *descendants* with a better *fitness* value than their parents.

Most engineering optimization design problems are difficult to solve using conventional algorithms since they comprise problem-specific constraints (linear, non-linear, equality or inequality). Despite the success of GA in a wide-range of applications, solving constrained optimization problems is no easy task. The most common technique is to apply penalty functions. As a result, the problem is converted from a constrained to an unconstrained optimization problem. The major drawback of these penalty functions is the requirement of a definition and proper tuning of their parameters, which can be challenging and problematic.

Hence, the aim of this paper is to answer one of the most stimulating questions encountered in meta-heuristics: constraint-handling in evolutionary algorithms. In this paper, we will use a GA as a numerical tool to propose a constraint-handling technique that eliminates the use of penalty functions. We present a parameter-free constraint-handling technique for GA using the violation factor; hence, the method will be referred to as VCH (Violation Constraint-Handling). This paper is organized as follows: Section 2 contains the most relevant constraint-handling techniques proposed by previous works. Then, the VCH approach is described in Section 3 and validated with the numerical examples in Section 4. Our technique and results are discussed in Section 5 and the conclusion and some paths for future research are provided in Section 6.

# Literature Review

Below are the most relevant constraint-handling techniques used in EA's for the purpose of this study. The reader is referred to the following surveys (Coello, 2002; Coello and Carlos, 1999; Dasgupta and Michalewicz, 1997; Gen and Cheng, 1996; Michalewicz, 1995a; 1995b; Michalewicz and Schoenauer, 1996; Yeniay, 2005) for further details, explanations and comparison.

*Penalty Methods*

The penalty methods are the most common approaches for constraint-handling in EA. Penalty functions were initially suggested by (Courant, 1943) and later extended by (Carroll, 1961) and (Fiacco and McCormick, 1966). Generally, the penalty term is determined from the amount of constraint violation of the solution vector. The formulation of the exterior penalty functions can be expressed as:

$$\psi(\vec{x}) = f(\vec{x}) \pm \left[ \sum_{i=1}^{n} a_i \times G_i + \sum_{j=1}^{m} b_j \times H_j \right] \quad (1)$$

where, $\psi(\vec{x})$ is the new fitness function to be optimized, $G_i$ and $H_j$ depend on the inequality constraints $g_i(\vec{x})$ and equality constraints $h_j(\vec{x})$ respectively and $a_i$, $b_j$ are called penalty factors.

The determination of the magnitude of the penalty term is a vital concern. The penalty term cannot be too high or else the algorithm will be locked inside the feasible domain and cannot move towards the border with the infeasible area. Too low, the term will be irrelevant in regard to the objective function and the search will remain in the infeasible region. Knowing how to exploit the search space in order to guide the search in the utmost desired direction is still unclear and rather challenging.

*Static Penalty*

In this group, the penalty factors remain constant during the evolution process and do not vary during each generation. A popular method is to define several levels of violation and attribute to each higher level a greater penalty coefficient $A_{ki}$. Homaifar *et al.* (1994) proposed to convert the equality constraints into inequality constraints and evaluate the following:

$$\psi(\vec{x}) = f(\vec{x}) + \sum_{i=1}^{m+n} (A_{k,i} \times \max[0, g_i(\vec{x})]^2) \quad (2)$$

Other researchers (Hoffmeister and Sprave, 1996; Morales and Quezada, 1998) have proposed interesting static penalties, but the main downside in these approaches are the necessity of a high number of parameters. They are difficult to describe and may not always be easy to obtain for real-world applications.

*Dynamic Penalty*

In this category, the penalty function depends on the generation number and usually the penalty term will increase over each generation. Joines and Houck (1994) evaluate each individuals using the following expressions:

$$\psi(\vec{x}) = f(\vec{x}) + (0.5 \times t)^\alpha \times SVC(\beta, \vec{x})$$
$$SVC(\beta, \vec{x}) = \sum_{i=1}^{n} A_i^\beta(\vec{x}) + \sum_{j=1}^{m} B_j(\vec{x}) \quad (3)$$

and:

$$A_i(\vec{x}) = \begin{cases} 0, & \text{if } g_i(\vec{x}) \leq 0 \\ |g_i(\vec{x})|, & \text{otherwise} \end{cases}$$

$$B_j(\vec{x}) = \begin{cases} 0 & \text{if } -\epsilon \leq h_j(\vec{x}) \leq \epsilon \\ |h_j(\vec{x})|, & \text{otherwise} \end{cases}$$

The cooling parameters $\alpha$ and $\beta$ are user defined constants, $g_i$ and $h_j$ are the inequality and equality constraints respectively.

A common dynamic penalty function is based on the notion of simulated annealing (Kirkpatrick and Vecchi, 1983; Michalewicz and Attia, 1994), where the penalty term is updated on every occasion the solution is locked in near a local optimal. Dynamic penalties that learn from the search process are called adaptive penalty functions.

An incorrect choice of the penalty factor may lead to a local feasible solution or an infeasible solution (Back *et al.*, 1997). In regards to the simulated annealing, the solution is extremely sensitive to the cooling parameters.

*Co-Evolution*

Coello (2000b; 1999) proposed to evaluate the following fitness function with only inequality constraints as follows:

$$\psi(\vec{x}) = f(\vec{x}) - (CV \times w_1 + Viol \times w_2) \quad (4)$$

with $w_1$ and $w_2$ two integers considered as penalty factors, *Viol* is an integer that is incremented for each violated constraint and *CV* is the sum of all violated constraints expressed as:

$$CV = \sum_{i=1}^{n} \max[0, g_i(\vec{x})] \quad (5)$$

The idea of this method is to use a population to evolve the solution vector and another to develop the penalty factors $w_1$ and $w_2$. This technique still requires the definition of four parameters and according to the author, they must be empirically determined. A major drawback of this penalty method is that it is very subtle to variations in the parameters in addition of their rigorous definition and high computational cost.

*Death Penalty*

A major concern in optimization algorithms in general and in EA's in particular is the element of 'infeasible solutions'. The simplest way is to reject the individual-(hence 'death') when at least one constraint is violated. A new point is generated until a feasible solution is found, therefore making this approach a lengthy process with the high risk of stagnating.

*Separation of Objectives and Constraints*

There are more than a few proposed approaches that separate the amount of constraint violation and the objective function. For instance, Powell and Skolnick (1993) scale the objective function $f(\vec{x})$ into the interval $]-\infty, 1]$, whereas $g_i(\vec{x})$ and $h_j(\vec{x})$ are scaled into the interval $[1, +\infty[$ and when the solution is unfeasible the objective function is not combined with the penalty term. During the search, each individual is assessed according to the following form:

$$\psi(\vec{x}) = \begin{cases} f(\vec{x}) \text{ if feasible} \\ 1 + A\left(\sum_{i=1}^{n} g_i(\vec{x}) + \sum_{j=1}^{m} h_j(\vec{x})\right) \text{otherwise} \end{cases} \quad (6)$$

with *A* a constant to be determined by the user.

The main difficulty with Powell and Skolnick (1993) is not the definition of the penalty factor *A* but rather with the concept of superiority of feasible over infeasible solutions. Deb (2000) uses a similar separation approach and evaluates the individuals using:

$$\psi(\vec{x}) = \begin{cases} f(\vec{x}) \text{ if feasible} \\ f_{worst} + \sum_{i=1}^{m+n} k_i(\vec{x}) \text{otherwise} \end{cases} \quad (7)$$

where, $f_{worst}$ is the worst feasible solution in the population and $k_i(\vec{x})$ include the inequality constraints and the transformed equalities. The constraints are normalized since they are each expressed in different units and to avoid any preference.

**Proposed Technique**

One of the key complications in using GA for practical engineering optimization applications is the design of the fitness function. When dealing with constrained problems, we must find a mean to estimate the closeness of an infeasible solution to the feasible region. By simply examining the previously proposed constraint-handling techniques, several key points can be derived about the existing methods. Initially, they are diverse, yet require the definition and fine-tuning of at least one parameter.

Apart of being an arduous procedure to define and control the penalty terms, we claim that such methods deviate from the essence of the philosophy of the evolutionary algorithms (i.e., techniques based on the principle of natural selection). Arguably, the most widely used algorithm is the genetic algorithm developed by (Holland, 1975). Despite the success of GA's as optimization techniques in many engineering applications, they are mostly applied on unconstrained problems.

Therefore, the main proposal of the authors is to suggest a constraint-handling technique that preserves the notions of the GA. The key motif is to keep the fitness function equivalent to the designer's objective and eliminate any additional penalty functions. The core structure of GA is analogous to the theory of biological evolution mimicking the principle of the survival of the fittest. The proposed constraint-handling technique is directly inspired from the nature of genetic algorithms, since the objective function is preserved during the evolution process. In this study, we will implement the proposed VCH method inside a genetic algorithm due to its advantages:

a) **Adaptability**: Does not oblige the objective function to be continuous or in algebraic form.
b) **Robustness**: Escapes more easily from local optimums because of its population-based nature.
c) **Equilibrium**: Provide a good balance between exploitation and exploration. Do not need specific domain information, but can be further exploited it if provided.
d) **Flexibility**: GA's are simple and relatively easy to implement.

We are interested in the general nonlinear programming problems (NLP); a minimization or maximization of a constrained optimization problem in which we want to:

$$\text{Find } \vec{x} \text{ which minimizes } f(\vec{x}) \qquad (8)$$

Subject to certain set of constraints:

$$g_i(\vec{x}) \leq 0, i = 1,\ldots,n$$
$$h_j(\vec{x}) = 0, j = 1,\ldots,m$$
$$\vec{x}_k^L \leq \vec{x}_k \leq \vec{x}_k^U, k = 1,\ldots,p$$

where, $\vec{x}$ is the solution vector with $p$ variables $\vec{x} = [x_1, x_2, \ldots x_p]$, $n$ is the number of inequality constraints, $m$ the number of equality constraints and the $k^{th}$ variable varies in the range $[\vec{x}_k^L, \vec{x}_k^U]$; the lower and upper bounds for each variable.

These constraints can be either linear or non-linear. Most constraint-handling approaches tend to deal with inequality constraints only. Therefore, a customary approach is to transform equality to inequality constraints using the following expression:

$$|h_j(\vec{x})| - \in \leq 0 \qquad (9)$$

which is equivalent to $h_j(\vec{x}) - \in \leq 0$ and $-h_j(\vec{x}) - \in \leq 0$, where $\in$ is the tolerance (usually a very small value, user-defined). This is justified by the fact that obtaining sampling points that satisfy the equality exactly is very difficult and hence, some tolerance or allowance is used in practice.

We shall first illustrate the overall procedure of the VCH technique for GA. In the subsequent, we assume the following:

- *PopNum*: Population length
- *Nelite*: Number of elites
- *Ncross*: Number of crossover-ed individuals
- *Nmut*: Number of mutated individuals (Bmut = PopNum-Nelite-Ncross)

- Real-coded GA according to which each chromosome is a string of the form ⟨$d_1,d_2,\ldots,d_m$⟩, where $d_1, d_2 \ldots, d_m$ are real numbers

**Step 1:** Initialization of the population:

The design variables are randomly initialized to satisfy the upper and lower constraints as follows:

$$\vec{x} = \vec{x}_k^L + [\vec{x}_k^U - \vec{x}_k^L] * rand(0,1) \qquad (10)$$

**Step 2:** Evaluation of the fitness function, normalized constraints and constraint violation:

For each individual $\vec{x}$, the fitness function $f(\vec{x})$ is calculated along with the resulting constraints. All the equality constraints are converted into inequalities using (9), hence a total of *n+m* inequality constraints. These equations are all normalized and therefore become in the form of:

$$G_i = \text{normalized } g_i(\vec{x}), i = 1,\ldots,n \qquad (11)$$

$$G_i = h_j(\vec{x})/\in -1 \leq 0, \ j = n+1,\ldots n+m \qquad (12)$$

Furthermore, the amount of Constraint Violation (C.V) of the normalized constraints $G_k$, ($k = 1, \ldots n+m$), is determined using:

$$C.V = \sum_{i=1}^{i=n+m} \max(0, G_i) \qquad (13)$$

In addition, the number of violation is defined as the percentage of violated constraints for a given solution:

$$N.V = \frac{\text{number of violated constraints}}{n+m} \qquad (14)$$

**Step 3:** Sorting of the population:

The population is separated into two families; feasible solutions ($V_0$) and unfeasible ($V_1$) consisting of individuals that violate at least one constraint. The first set ($V_0$) is sorted with respect to the fitness value (ascending order, assuming a minimization problem). The second family ($V_1$) is sorted according to the proposed pair-wise comparison rules. In the VCH approach, we adopted a feasibility-based rule, a set of rules to evolve the population at each generation:

- If one individual is infeasible and the other is feasible, the winner is the feasible solution
- If both individuals are feasible, the winner is the one with the highest fitness value
- If both individuals are infeasible, the winner is the one with the lowest Number of Violations (N.V)
- If both individuals are infeasible with the same (N.V), the one with the lowest Constraints Violation (C.V) value wins

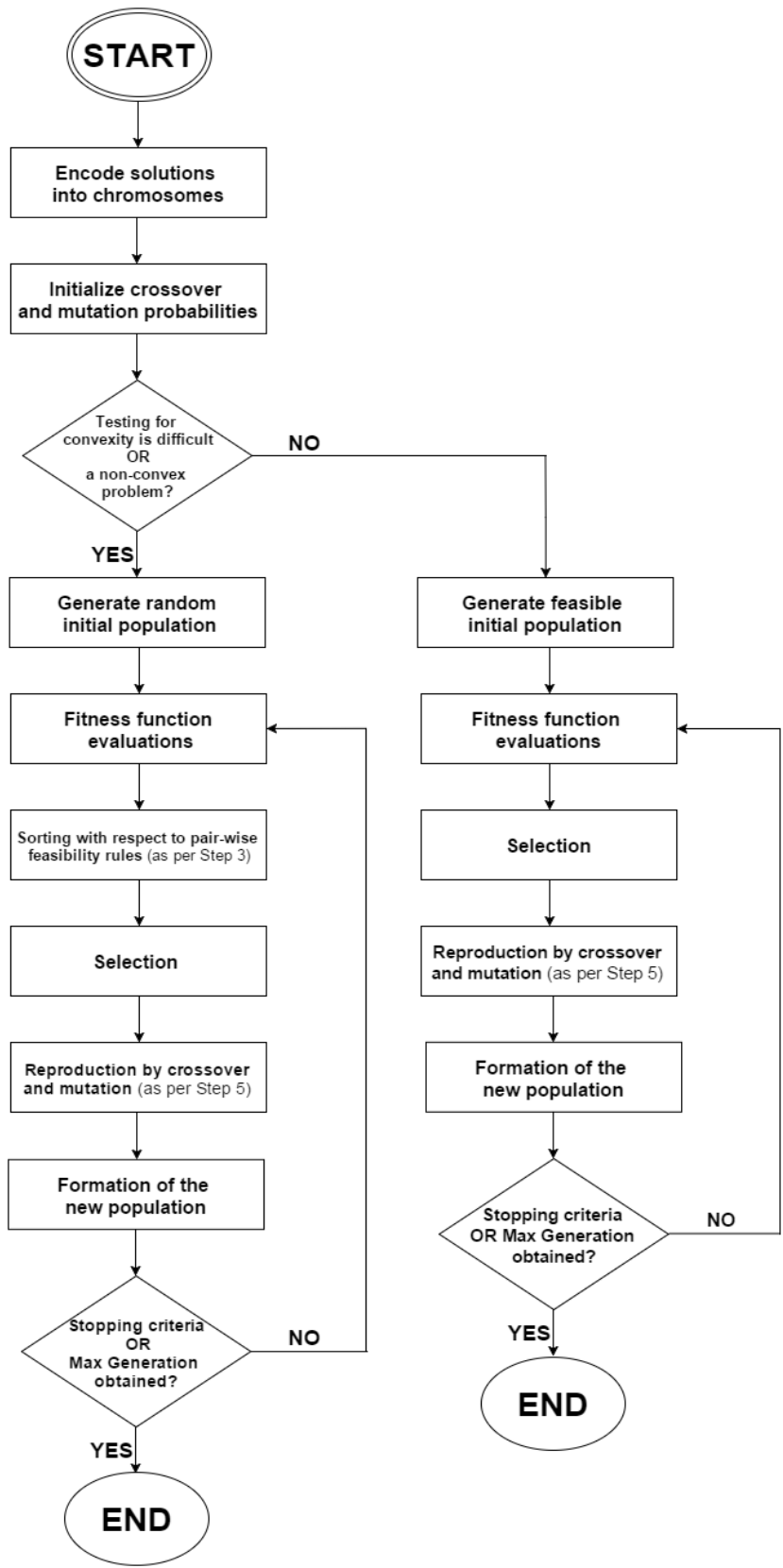

Fig. 1. Complete flowchart of the proposed GA

**Step 4:** Formation of Elites:

The sorted families $V_0$ and $V_1$ form the new population. The first *Nelite* individuals are the elites, which are kept intact to the next generation with no alteration. This selection operator, one form of elitism consists of a driving force for self-organization or convergence and is essentially an intensive exploitation.

**Step 5:** Reproduction by crossover and mutation:

A tournament-based technique is used to perform the crossover among the individuals of the population. Whole arithmetic crossover (Michalewicz and Janikow, 1996; Michalewicz and Nazhiyath, 1995; Michalewicz and Schoenauer, 1996) is applied in our algorithm. It is composed of a linear combination of two parent vectors to be crossed (as shown in 15). This genetic operator uses simple static parameter $\phi$ (a random number between 0 and 1). Any linear combination of two feasible points in a convex domain will produce another feasible point (Michalewicz, 1992).

$$A \otimes B = A * \varnothing + B * (1 - \varnothing) \quad (15)$$
with $\phi = rand(0,1)$

The reproduction and crossover operators are programmed to imitate the paradigm of the survival of the fittest. The crossover operator is a recombination of two chromosomes, an operation that ensures an efficient exploitation in the local search within a subspace. Therefore, the offspring are spread over the entire feasible space. The crossover-elitism pair eases the achievement of global optimality. In contrast, the mutation operator is a randomization mechanism for global search and exploration.

**Step 6:** Stopping criteria:

Steps 2-5 are repeated until either the stopping criteria is respected or the maximum number of generation is attained. We implemented a severe stopping criterion on the best solution of each generation; the relative error between the present and the past generation for each design variable must remain less than the user-defined tolerance for at least *N* amount of generations.

As the population evolves, the proposed VCH process will lead the search to reach feasible regions, much similar to a severe penalty function. Nonetheless, in order to maintain infeasible solutions near the feasible region, at each generation, the infeasible solution with the lowest C.V and best objective function value will be kept in the population for the next generation. As a result, the population will most likely have fewer infeasible solutions located in promising areas of the search space. The VCH approach does not use any penalty function to handle the constraints. Instead, it can be seen to have a mechanism that encourages the solutions close to the feasible region in favorable areas of the design space to remain in the population. This does not add substantial computational cost.

An optimization problem is called a convex programming problem if the objective function and the constraint functions are both convex. Originally, EAs were developed to solve unconstrained problems. Constrained optimization is a computationally challenging task, mainly if the constraint functions are nonlinear and/or nonconvex. A positive feature of the proposed VCH approach is that it does not care about the structure of the constraint functions (linear or nonlinear, convex or nonconvex). An accelerated VCH technique for convex optimization problems is to generate an initial population with only feasible solutions. Thereafter, the reproduction by means of an arithmetic crossover (as per expression 15) will continue to generate feasible solutions (Michalewicz, 1992). Testing for convexity or concavity can be done by evaluating if the Hessian matrix $H(X) = \left[ \dfrac{\partial^2 f(X)}{\partial x_i \partial x_j} \right]$ is positive semi definite (for minimization problems). The accelerated genetic algorithm for the solving of constrained problems in the case of convex design and objective spaces would not require the use of any feasibility-rules. Rather, solutions with high fitness values are preferred since all the individuals of the population are feasible (as described in Fig. 1).

## Numerical Examples

In order to validate the proposed constraint handling technique, several examples taken from the literature will be used. These numerical examples are all constrained optimization problems that include linear and nonlinear constraints. These are benchmark optimization problems that have been previously evaluated by other GA-based techniques, which is useful to investigate and demonstrate the quality and usefulness of the proposed VCH approach.

The algorithm is implemented in Matlab (R2013 a Student Version 8.1.1.604) run by a 2.90 GHz Intel® Core™ i7-3520M CPU (4 Duo processor) with 4096 MB of Random Access Memory (RAM). The number of crossover-ed and mutated individuals in the population (100 chromosomes) are 94 and 5 respectively.

That means only one individual is preserved to the following generation based on elitism. The termination criterion is taken as either the reach of the maximum number of generations (set to 500 in all examples) or the achievement of the relative error on the design vector (set to be equal to $10^{-6}$). To demonstrate the effectiveness of the proposed VCH, the best, mean, median, worst and fitness evaluations are recorded for 20 independent runs. We are concerned with the efficiency of the technique in terms of CPU time, because we are particularly interested in solving engineering optimization problems, for which the cost of fitness evaluations is generally high. However, it is more convenient to adopt the number of fitness evaluations since it is independent of the implemented hardware. The stopping condition employed in the numerical simulations is identical to the criteria described in step 6.

*Himmelblau's Nonlinear Optimization Problem*

This problem was originally proposed by (Himmelblau, 1972) and has been widely used as a point of reference for nonlinear constrained optimization problems and several other constraint handling techniques that use penalties. In this formulation, there are five design variables [$x_1$, $x_2$, $x_3$, $x_4$, $x_5$], six nonlinear inequality constraints and 10 boundary conditions.

The problem can be stated as follows:

$$\text{Minimize} f(\vec{x}) = 5.3578547$$
$$x_3^2 + 0.8356891\, x_1 x_5 + 37.293239\, x_1 - 40792.141$$

Subject to:

$$g_1(\vec{x}) = 85.334407 + 0.0056858\, x_2 x_5$$
$$+ 0.00026\, x_1 x_4 - 0.0022053\, x_3 x_5 - 0.0022053\, x_3 x_5$$
$$g_2(\vec{x}) = 80.51249 + 0.0071317\, x_2 x_5$$
$$+ 0.0029955\, x_1 x_2 + 0.0021813\, x_3^2 + 0.0021813 x_3^2$$
$$g_3(\vec{x}) = 9.300961 + 0.0047026\, x_3 x_5$$
$$+ 0.0012547\, x_1 x_3 + 0.0019085\, x_3 x_4$$
$$0 \leq g_1(\vec{x}) \leq 92 \quad 90 \leq g_2(\vec{x}) \leq 110 \quad 20 \leq g_3(\vec{x}) \leq 25$$

*Himmelblau's Nonlinear Problem*

The best solution was found to be $f(\vec{x}) = -30988.951$, with 15000 evaluations only. The design vector is: $x_1 = 78.0029$, $x_2 = 33.080$, $x_3 = 27.353$, $x_4 = 44.61$ and $x_5 = 44.264$. The mean is $f(\vec{x}) = -30845.42278$, with a standard deviation of 48.60 (as listed in Table 2). The worst solution found was $f(\vec{x}) = -30800.89145$, which is better than 75% of the reviewed methods as per Table 1. The significantly fewer function evaluations reduced the computational cost of the optimization procedure to an average CPU time of 0.52 s/run for 20 independent runs.

*Minimization of the Weight of a Tension/Compression Spring*

This optimization problem was described by (Arora, 1989) and (Belegundu, 1983) and it consists of minimizing the weight of a tension/compression spring, subject to constraints on minimum deflection, shear stress, surge frequency, outside diameter and on the design variables. The later are the wire diameter $d$ ($= x_1$), the mean coil diameter $D$ ($= x_2$) and the number of active coils $N$ ($= x_3$).

The problem is expressed as follows:

$$\text{Minimize } f(\vec{x}) = (x_3 + 2)x_1^2 x_2$$

Subject to:

$$g_1(\vec{x}) = 1 - \frac{x_2^3 x_3}{71785 x_1^4} \leq 0$$
$$g_3(\vec{x}) = 1 - \frac{140.45 x_1}{x_2^2 x_3} \leq 0 \quad g_4(\vec{x}) = \frac{x_1 + x_2}{1.5} - 1 \leq 0$$
$$g_2(\vec{x}) = \frac{4 x_2^2 - x_1 x_2}{12566(x_1^3 x_2 - x_1^4)} + \frac{1}{5108 x_1^2} - 1 \leq 0$$

*Spring Design*

The optimal solution for this problem is at: $x_1 = 0.0513412$, $x_2 = 0.3483225$, $x_3 = 11.80261$, with an optimal fitness value of $f(\vec{x}) = 0.012672$, obtained after 28000 evaluations (as per Table 3 and 4). The mean is 0.01269293 with a low standard deviation of $8.32845 \times 10^{-6}$.

Table 1. Optimal results for Himmelblau's nonlinear problem (NA = Not Available)

| Method | Design variables | | | | | |
| --- | --- | --- | --- | --- | --- | --- |
| | $x_1$ | $x_2$ | $x_3$ | $x_4$ | $x_5$ | $f(\vec{x})$ |
| Present study | 78.00 | 33.08 | 27.35 | 44.61 | 44.26 | -30988.951 |
| Coello (2000a) | 78.59 | 33.01 | 27.64 | 45.00 | 45.00 | -30810.359 |
| Deb (1997) | NA | NA | NA | NA | NA | -30665.539 |
| Deb (2000) | 78.00 | 33.00 | 29.99 | 45.00 | 36.77 | -30665.5 |
| Homaifar *et al.* (1994) | NA | NA | NA | NA | NA | -30575.86 |
| (Bean, 1994; Ben Hadj-Alouane and Bean, 1997) | NA | NA | NA | NA | NA | -30560.361 |
| Gen and Cheng (1997) | 81.49 | 34.09 | 31.24 | 42.20 | 34.37 | -30183.575 |
| Coello and Cortés (2004) | NA | NA | NA | NA | NA | -30665.51 |

Table 2. Statistical results for Himmelblau's nonlinear problem (NA = Not Available)

| Method | Mean | Worst | Std | Fitness Evaluations |
|---|---|---|---|---|
| Present study | -30845.422 | -30800.891 | 48.60797 | 15000 |
| Coello (2000b; 1999) | -30984.240 | -30792.407 | 73.63353 | 900000 |
| Coello (2000a) | NA | NA | NA | 16000 |
| Deb (2000) | -30665.53 | -29846.65 | NA | 250000 |
| Homaifar *et al.* (1994) | -30403.87 | -30294.50 | 64.19172 | 40000 |
| (Bean, 1994; Ben Hadj-Alouane and Bean, 1997) | -30397.40 | -30255.37 | -73.8032 | NA |
| Gen and Cheng (1997) | NA | NA | NA | NA |
| Coello and Cortés (2004) | -30654.98 | -30517.44 | 32.67 | 150000 |

Table 3. Optimal results for tension/compression spring design problem (NA = Not Available)

| | Design variables | | | |
|---|---|---|---|---|
| Method | $x_1$ | $x_2$ | $x_3$ | $f(\vec{x})$ |
| Present study | 0.0513412 | 0.3483225 | 11.80261 | 0.012672 |
| Coello and Mezura-Montes (2002) | 0.051989 | 0.363965 | 10.890522 | 0.012681 |
| Mezura-Montes and Coello (2005) | 0.052836 | 0.384942 | 9.807729 | 0.012689 |
| Coello (2000b; 1999) | 0.05148 | 0.351661 | 11.632201 | 0.012704 |

Table 4. Statistical results for tension/compression spring design problem (NA = Not Available)

| Method | Mean | Worst | Std | Fitness Evaluations |
|---|---|---|---|---|
| Present study | 0.0126929 | 0.01270562 | 8.32845E-06 | 28000 |
| Coello and Mezura-Montes (2002) | 0.012742 | 0.012973 | 5.9E-05 | 80000 |
| Mezura-Montes and Coello (2005) | 0.013165 | 0.014078 | 3.90E-04 | 30000 |
| Coello (2000b; 1999) | 0.0127692 | 0.01282082 | 3.939E-05 | 900000 |

Table 5. Optimal results for pressure vessel design problem (NA = Not Available)

| | Design variables | | | | |
|---|---|---|---|---|---|
| Method | $x_1$ | $x_2$ | $x_3$ | $x_4$ | $f(\vec{x})$ |
| Present study | 0.8125 | 0.4375 | 42.0978 | 176.644 | 6059.79164 |
| Mezura-Montes and Coello (2005) | 0.8125 | 0.4375 | 42.0984 | 176.636 | 6059.7143 |
| Coello and Mezura-Montes (2002) | 0.8125 | 0.4375 | 42.0974 | 176.654 | 6059.946 |
| Coello and Cortes (2004) | 0.8125 | 0.4375 | 42.0869 | 176.779 | 6061.1229 |
| Coello (2000c) | 0.875 | 0.5000 | 42.0939 | 177.080 | 6069.3267 |
| Coello (2000b; 1999) | 0.8125 | 0.4375 | 40.3239 | 200.00 | 6288.7445 |
| Deb (1997) | 0.9375 | 0.5000 | 48.3290 | 112.679 | 6410.3811 |
| Yun (2005) | 1.125 | 0.6250 | 58.2850 | 43.725 | 7198.424 |
| Wu and Chow (1995) | 1.125 | 0.6250 | 58.1978 | 44.293 | 7207.494 |

## Design of a Pressure Vessel

This problem was originally proposed (Sandgren, 1988; 1990) for the design of a pressure vessel with minimal overall cost (material, forming and welding). The air storage tank has a working pressure of 2000 psi and a maximum volume of 750 ft$^3$. There are four design variables namely; $T_S (= x_1)$ thickness of the shell, $T_h (= x_2)$ thickness of the head, $R (= x_3)$ inner radius and $L (= x_4)$ length of the cylindrical section of the vessel, not including the head. $T_S$ and $T_h$ are integer multiples of 0.0625 inch and $R$ and $L$ are continuous.

The following pressure vessel design problem is taken from Kannan and Kramer (1994) as follows:

Minimize $f(\vec{x}) = 0.6224 x_1 x_3 x_4$
$+ 1.7781 x_2 x_3^2 + 3.1661 x_1 x_4^2 + 19.84 x_1 x_3^2$

Subject to:

$g_1(\vec{x}) = -x_1 + 0.0193 x_3 \leq 0 \qquad g_4(\vec{x}) = x_4 - 240 \leq 0$
$g_2(\vec{x}) = -x_2 + 0.00954 x_3 \leq 0$
$g_3(\vec{x}) = -\pi x_3^2 x_4 - \frac{4}{3}\pi x_3^3 + 1,296,000 \leq 0$

## Welded Beam Design Problem

The welded beam problem has been used as a benchmark problem originally proposed by (Rao, 1996).

Table 6. Statistical results for pressure vessel design problem (NA = Not Available)

| Method | Mean | Worst | Std | Fitness Evaluations |
|---|---|---|---|---|
| Present study | 6060.06181 | 6060.21499 | 0.12847724 | 24250 |
| Mezura-Montes and Coello (2005) | 6379.938037 | 6820.397461 | 2.10E+02 | 30000 |
| Coello and Mezura-Montes (2002) | 6177.253268 | 6469.32201 | 130.929702 | 80000 |
| Coello and Cortes (2004) | 6734.0848 | 7368.0602 | 457.9959 | 150000 |
| Coello (2000) | 6177.253268 | 6469.32201 | 130.929702 | 50000 |
| Coello (2000b; 1999) | 6293.843232 | 6308.14965 | 7.41328537 | 900000 |

Table 7. Optimal results for welded beam design problem (NA = Not Available)

| Method | Design variables | | | | |
|---|---|---|---|---|---|
| | $x_1$ | $x_2$ | $x_3$ | $x_4$ | $f(\vec{x})$ |
| Present study | 0.20578 | 3.47294 | 9.02922 | 0.20608 | 1.726718 |
| Coello and Mezura-Montes (2002) | 0.20599 | 3.47133 | 9.02022 | 0.20648 | 1.728226 |
| Coello (2000b) | 0.20880 | 3.42050 | 8.99750 | 0.21000 | 1.7483094 |
| Siddall (1972) | 0.2444 | 6.2189 | 8.2915 | 0.2444 | 2.3815433 |
| Ragsdell and Philipps (1976) | NA | NA | NA | NA | 2.385937 |
| Deb (1991) | 0.2489 | 6.173 | 8.1789 | 0.2533 | 2.433116 |

Table 8. Statistical results for welded beam design problem (NA = Not Available)

| Method | Mean | Worst | Std | Fitness evaluations |
|---|---|---|---|---|
| Present study | 1.727529953 | 1.72807450 | 0.0004207 | 30000 |
| Mezura-Montes *et al.* (2007) | 1.725 | 1.725 | 1.00E-15 | 24000 |
| Coello and Mezura-Montes (2002) | 1.792654 | 1.993408 | 0.074713 | 80000 |
| Coello (2000b) | 1.77197269 | 1.78583465 | 0.0112228 | 900000 |

The beam is designed for minimum cost subject to constraints on shear stress ($\tau$), bending stress in the beam ($\sigma$), buckling load on the bar ($P_c$) end deflection of the beam ($\delta$) and side constraints. In this problem there are four design variables namely; thickness of the beam $h(=x_1)$, length of the welded joint $l(=x_2)$, width of the beam $t(=x_3)$ and thickness of the beam $b(=x_4)$. It is important to note that in this problem, there are several models in the overviewed literature, with different number of constraints and variable definitions. In the present study, the results for the following optimization formulation are presented:

Minimize $f(\vec{x}) = 1.1047x_1^2 x_2 + 0.04811 x_3 x_4 (14 + x_2)$

Subject to:

$g_1(\vec{x}) = \tau(\vec{x}) - \tau_{max} \leq 0$ $\quad$ $g_2(\vec{x}) = \sigma(\vec{x}) - \sigma_{max} \leq 0$
$g_3(\vec{x}) = x_1 - x_4 \leq 0$ $\quad$ $g_4(\vec{x}) = 0.125 - x_1 \leq 0$
$g_5(\vec{x}) = \delta(\vec{x}) - \delta_{max} \leq 0$ $\quad$ $g_6(\vec{x}) = P - P_c(\vec{x}) \leq 0$
$g_7(\vec{x}) = 0.10471 x_1^2 + 0.04811 x_3 x_4 (14.0 + x_2) - 5.0 \leq 0$

where, $\tau$ is the shear stress in the weld (it has two components namely primary stress $\tau'$ and secondary stress $\tau''$), $\tau_{max}$ is the allowable shear stress of the weld (= 13600 psi), $\sigma$ the normal stress in the beam, $\sigma_{max}$ is the allowable normal stress for the beam material (= 30000 psi), $P_c$ the buckling load, $P$ the load (= 6000 lb) and $\delta$ the beam end deflection:

$$\tau(\vec{x}) = \sqrt{(\tau')^2 + \frac{2\tau'\tau''x_2}{2R} + (\tau'')^2}, \tau' = \frac{P}{\sqrt{2}x_1 x_2}, \tau'' = \frac{MR}{J}$$

$$M = P\left(L + \frac{x_2}{2}\right), R = \sqrt{\frac{x_2^2}{4} + \left(\frac{x_1 + x_3}{2}\right)^2}, \sigma(\vec{x}) = \frac{6PL}{x_4 x_3^2}$$

$$J = 2\left\{\sqrt{2}x_1 x_2 \left[\frac{x_2^2}{12} + \left(\frac{x_1 + x_3}{2}\right)^2\right]\right\}, \delta(\vec{x}) = \frac{4PL^3}{Ex_3^3 x_4},$$

$$P_c(\vec{x}) = \frac{4.013E\sqrt{x_3^2 x_4^6/36}}{L^2}\left(1 - \frac{x_3}{2L}\sqrt{\frac{E}{4G}}\right)$$

$L = 14$ in $\quad \delta_{max} = 0.25$ in $\quad E = 30 \times 10^6$ psi $\quad G = 12 \times 10^6$ psi

*Welded Beam Design*

The presented algorithm has been tested on this optimization problem and compared with the best solutions by previous methods reported in Table 7. The optimal design vector was found to be: $x_1 = 0.20578$, $x_2 = 3.47294$, $x_3 = 9.02922$, $x_4 = 0.20608$ with an optimal fitness value $f(\vec{x}) = 1.726718$. In average, the time elapsed for one execution of the program is 1.82 sec and the average number of fitness evaluations for 20 runs is 30000.

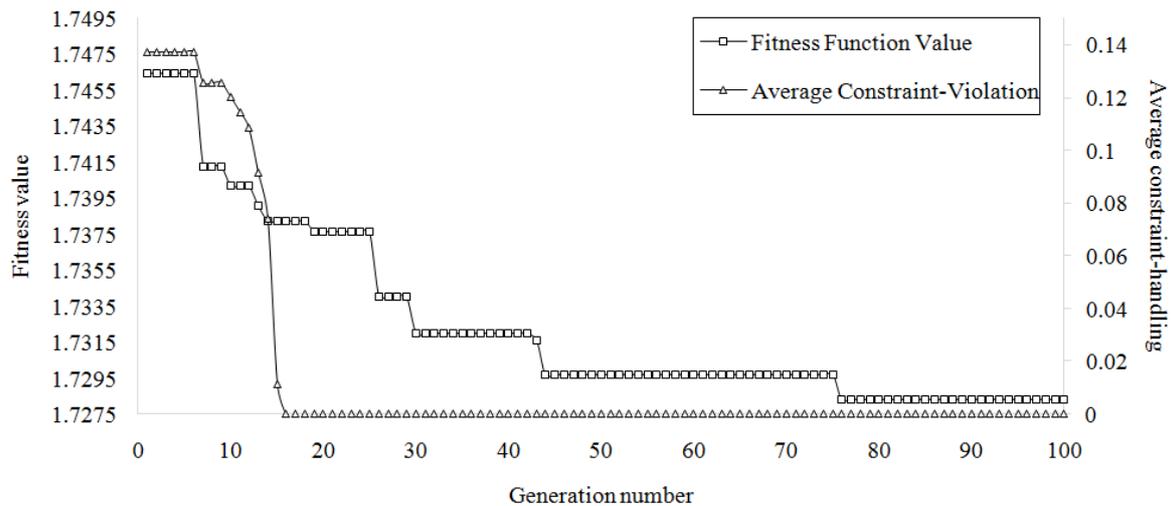

Fig. 2. Average constraint-handling (CV) and best fitness function obtained with the proposed VCH method for numerical example 4

## Discussion

In this paper, we have developed a constraint-handling method for GA, free of any penalty parameters using only the violation factor called VCH (as summarized in the flowchart of Fig. 1) that is capable of sorting a population with both feasible and infeasible individuals. In the proposed VCH method, at a given iteration, the individuals of the population are never compared in terms of both objective function value and constraint violation information. Essentially, the main motif is to keep the fitness function equivalent to the designer's objective function and therefore the conversion of the constrained problem into an unconstrained one is no longer required.

Genetic algorithms try to mimic the principle of the survival of the fittest, where newer generations are evolved in attempt to produce descendants with a better 'fitness'. Because at all times the fitness function is equal to the objective function to be minimized, our proposed VCH technique is more conforming with the biological fundamentals of genetic algorithms. A major drawback of many techniques in the literature is that the constraint handling method requires a feasible initial population. For some problems, finding a feasible solution is NP-hard and even impossible for the problems with conflicting constraints. In the VCH approach, it is not required to have a feasible initial population. There are available techniques that ensure feasibility of the population when dealing with linear constraints such as (Yun, 2005) by means of mathematical programming.

Compared to other constraint-handling techniques based on penalty functions, the VCH method was able to provide a consistent performance and demonstrated to be simpler, faster and delivered reliable optimal solutions without any violation of the constraints. As the population evolves, the VCH method will lead the search to reach faster feasible regions. This is revealed in Fig. 2, with the convergence of the average constraint violation of the elites towards zero (no violation) as the population evolves. The VCH method allows the closest solutions to the feasible region in favorable areas of the search space to remain in the population. Specific methods such as the reduced gradient method, cutting plane method and the gradient projection method are appropriate. However, they are only fitting either to problems having convex feasible regions or with few design variables. Furthermore, the overall results suggest that the proposed approach is highly competitive and was even able to contest (some cases improve) the results produced by other methods, some of which are more difficult constraint-handling techniques applied to genetic algorithms. The VCH algorithm was tested on several benchmark examples and demonstrated its ability to solve problems with a large number of constraints.

## Conclusion and Future Work

The diversity and popularity of evolutionary algorithms does not imply that there are no problems that need urgent attention. From one point of view, these optimization algorithms are very good at obtaining optimal solutions in a practical time. On the other, they still lack in balance of accuracy, computational efforts, global convergence and the tuning and control of their parameters. Nature has evolved over millions of years, providing a rich source of inspiration for researchers to develop diverse algorithms with different degrees of success and popularity. Such diversity and accomplishment does not signify that we should focus solely on developing more algorithms for the sake of algorithm development, or even worse for the sake of publication. This attitude distracts from the search for solutions for more challenging and truly important problems in optimization and new algorithms may be proposed only if they:

- deliver truly novel ideas
- demonstrate to be efficient techniques that solve challenging optimization problems (that are not solved by existing methods)
- verify to the "3-self" (self-adaptive, self-evolving and self-organizing algorithms)

It is vital to state that during the development of this technique, several other versions of the same approach were examined without much success. For example, different reproduction probabilities (crossover and mutation) were tested. The authors avoided a high mutation rate to prevent the method of becoming a random search, but choose to keep it at 0.05 for a more robust global search and exploration. More than a few other crossover techniques were examined but the objective of this paper was not to present a comparative study on their performance but rather present the parameter-free constraint-handling technique using the violation factor. It is still unclear how to achieve optimal balance of exploitation and exploration by proper parameter tuning of the evolutionary operators of GA in general and in the VCH algorithm in particular. The crossover operator ensured an efficient exploitation in the local search within a subspace and can provide good convergence in local subspace. The selection and mutation operators enabled the GA to have a higher ability for exploration. It could be argued that the VCH technique is competent because it does not require any fine-tuning of the GA, which is normally performed by trial and error and is time consuming. Finally, it is worth mentioning that for many of them, it is unclear if the authors implemented a stopping criterion or not. In our study however, a severe criterion was introduced based on the minimum relative error of the design variables. Moreover, the user–defined tolerance has to be respected for a number of generations before the execution is terminated.

The main challenges that still require further examination are: The proof of convergence of some EA, control and tuning of parameters, the solution of large scale applications (e.g., the traveling salesman problem) and finally tackling Nondeterministic Polynomial (NP)-hard problems. Solving these issues is becoming more imperative than ever before. Among these matters is the open question of constraint-handling in GA specifically to solve engineering optimization problems. The insights gained by the proposed VCH technique should have an impact on the manner constrained problems are solved.

Lastly, the authors suggest in the upcoming work, further numerical simulations could be investigated for more complex optimization problems. It would be motivating to explore the integration of VCH technique in other EA's such as Particle Swarm Optimization (PSO), ant Colony Optimization (ACO), Bee Colony Optimization (BCO) and Differential Evolution (DE). Parameter tuning of the evolutionary operators in GA is an active area of research and could be examined in future work. Present work is aimed at introducing the proposed constraint-handling technique in a multi objective platform for the optimization of the composite lay-out of wind turbine blades using a genetic algorithm as discussed in (Chehouri *et al.*, 2015).

## Acknowledgement

The authors would like to thank everyone who has contributed to the progress of the research. We thank the anonymous referees for their useful suggestions. Finally, the corresponding author would like thank his loving friends and family, particularly my brother Adel Chehouri and my girlfriend who I love dearly.

## Funding Information

The authors greatly acknowledge the Université du Québec à Chicoutimi (UQAC), Anti-Icing Materials International Laboratory (AMIL), MMC team, École Doctorale des Sciences et de la Technologie (EDST), Lebanese University (LU) and the Fonds de Recherche Nature et Technologie (FRQNT) for their continued financial support.

## Author's Contributions

**Adam Chehouri:** Main author and researcher of the manuscript. Developed the initial draft of the proposed constraint-handling technique and conducted the numerical simulations.

**Rafic Younes:** Main supervisor who organized the writing and structure of the paper and monitored the research.

**Jean Perron and Adrian Ilinca:** Research advisor, supervision and monitoring of the research.

## Ethics

The authors confirm that this manuscript has not been published elsewhere and that no ethical issues are involved.

## References

Arora, J.S., 1989. Introduction to Optimum Design. 1st Edn., McGraw-Mill, New York, ISBN-10: 007002460X, pp: 625.

Back, T., D.B. Fogel and Z. Michalewicz, 1997. Handbook of Evolutionary Computation. 1st Edn., Taylor and Francis, ISBN-10: 0750308958, pp: 1130.

Bean, J.C., 1994. Genetic algorithms and random keys for sequencing and optimization. ORSA J. Comput., 6: 154-160. DOI: 10.1287/ijoc.6.2.154

Belegundu, A.D., 1983. Study of mathematical programming methods for structural optimization. Dissertat. Abs. Int. Part B: Sci. Eng.

Ben Hadj-Alouane, A. and J.C. Bean, 1997. A genetic algorithm for the multiple-choice integer program. Operat. Res., 45: 92-101. DOI: 10.1287/opre.45.1.92

Carroll, C.W., 1961. The created response surface technique for optimizing nonlinear, restrained systems. Operat. Res., 9: 169-184.
DOI: 10.1287/opre.9.2.169

Chehouri, A., R. Younes, A. Ilinca and J. Perron, 2015. Review of performance optimization techniques applied to wind turbines. Applied Energy, 142: 361-388. DOI: 10.1016/j.apenergy.2014.12.043

Coello, C.A.C., 2000a. Treating constraints as objectives for single-objective evolutionary optimization. Eng. Optimizat., 32: 275-308.
DOI: 10.1080/03052150008941301

Coello, C.A.C. 2000b. Use of a self-adaptive penalty approach for engineering optimization problems. Comput. Ind., 41: 113-127.
DOI: 10.1016/S0166-3615(99)00046-9

Coello, C.A.C., 2000c. Constraint-handling using an evolutionary multiobjective optimization technique. Civil Eng. Syst., 17: 319-346.
DOI: 10.1080/02630250008970288

Coello, C.A.C., 1999. Self-adaptive penalties for GA-based optimization. Proceedings of the Congress on Evolutionary Computation, Jul. 06-09, IEEE Xplore Press, Washington, DC.

Coello, C.A.C. and A. Carlos, 1999. A survey of constraint handling techniques used with evolutionary algorithms. Laboratorio Nacional de Informática Avanzada.

Coello, C.A.C. and E.M. Montes, 2002. Constraint-handling in genetic algorithms through the use of dominance-based tournament selection. Adv. Eng. Inform., 16: 193-203.
DOI: 10.1016/S1474-0346(02)00011-3

Coello, C.A.C. and N.C. Cortés, 2004. Hybridizing a genetic algorithm with an artificial immune system for global optimization. Eng. Optimizat., 36: 607-634. DOI: 10.1080/03052150410001704845

Coello, C.A.C., 2002. Theoretical and numerical constraint-handling techniques used with evolutionary algorithms: A survey of the state of the art. Comput. Methods Applied Mechan. Eng., 191: 1245-1287. DOI: 10.1016/S0045-7825(01)00323-1

Courant, R., 1943. Variational methods for the solution of problems of equilibrium and vibrations. Bull. Amer. Math. Soc, 49: 1-23.
DOI: 10.1090/S0002-9904-1943-07818-4

Darwin, C. and W.F. Bynum, 2009. The origin of species by means of natural selection: Or, the preservation of favored races in the struggle for life.

Dasgupta, D. and Z. Michalewicz, 1997. Evolutionary Algorithms in Engineering Applications. 1st Edn., Springer Science and Business Media, Berlin, ISBN-10: 3540620214, pp: 554.

Deb, K., 1997. Geneas: A Robust Optimal Design Technique for Mechanical Component Design. Evolutionary Algorithms in Engineering Applications, Dipankar Dasgupta, Zbigniew Michalewicz (Eds.), Springer Science and Business Media, Berlin, ISBN-10: 3540620214, pp: 497-514.

Deb, K., 1991. Optimal design of a welded beam via genetic algorithms. AIAA J., 29: 2013-2015.
DOI: 10.2514/3.10834

Deb, K., 2000. An efficient constraint handling method for genetic algorithms. Comput. Methods Applied Mechan. Eng., 186: 311-338.
DOI: 10.1016/S0045-7825(99)00389-8

Dorigo, M., V. Maniezzo and A. Colorni, 1996. Ant system: Optimization by a colony of cooperating agents. IEEE Trans. Syst. Man Cybernet. Part B, 26: 29-41. DOI: 10.1109/3477.484436

Eberhart, R.C. and J. Kennedy, 1995. A new optimizer using particle swarm theory. Proceedings of the 6th International Symposium on Micro Machine and Human Science, Oct. 4-6, IEEE Xplore Press, Nagoya, pp: 39-43. DOI: 10.1109/MHS.1995.494215

Fiacco, A.V. and G.P. McCormick, 1966. Extensions of SUMT for nonlinear programming: Equality constraints and extrapolation. Manage. Sci., 12: 816-828. DOI: 10.1287/mnsc.12.11.816

Gen, M. and R. Cheng, 1996. A survey of penalty techniques in genetic algorithms. Proceedings of IEEE International Conference on Evolutionary Computation, May 20-22, IEEE Xplore Prss, Nagoya, pp: 804-809. DOI: 10.1109/ICEC.1996.542704

Gen, M. and R. Cheng, 1997. Genetic Algorithms and Engineering Design. 1st Edn., John Wily and Sons, New York.

Glover, F., 1977. Heuristics for integer programming using surrogate constraints. Decis. Sci., 8: 156-166. DOI: 10.1111/j.1540-5915.1977.tb01074.x

Himmelblau, D.M., 1972. Applied Nonlinear Programming. 1st Edn., McGraw-Hill, New York, pp: 498.

Hoffmeister, F. and J. Sprave, 1996. Problem-independent handling of constraints by use of metric penalty functions.

Holland, J.H., 1975. Adaptation in natural and artificial systems: An introductory analysis with applications to biology, control and artificial intelligence.

Homaifar, A., C.X. Qi and S.H. Lai, 1994. Constrained optimization via genetic algorithms. Simulation, 62: 242-253. DOI: 10.1177/003754979406200405

Joines, J.A. and C.R. Houck, 1994. On the use of non-stationary penalty functions to solve nonlinear constrained optimization problems with GA's. Proceedings of the 1st IEEE Conference on Evolutionary Computation, Jun. 27-29, IEEE Xplore Press, Orlando, FL, pp: 579-584. DOI: 10.1109/ICEC.1994.349995

Kannan, B. and S.N. Kramer, 1994. An augmented Lagrange multiplier based method for mixed integer discrete continuous optimization and its applications to mechanical design. J. Mechanical Des., 116: 405-411. DOI: 10.1115/1.2919393

Kirkpatrick, S. and M. Vecchi, 1983. Optimization by simulated annealing. Science, 220: 671-680. DOI: 10.1126/science.220.4598.671

Mezura-Montes, E. and C.A.C. Coello, 2005. Useful infeasible solutions in engineering optimization with evolutionary algorithms. Proceedings of the 4th Mexican International Conference on Artificial Intelligence, Nov. 14-18, Springer, pp: 652-662. DOI: 10.1007/11579427_66

Mezura-Montes, E., C.A. Coello, J. Velázquez-Reyes and L. Muñoz-Dávila, 2007. Multiple trial vectors in differential evolution for engineering design. Eng. Optimizat., 39: 567-589. DOI: 10.1080/03052150701364022

Michalewicz, Z. and C.Z. Janikow, 1996. GENOCOP: A genetic algorithm for numerical optimization problems with linear constraints. Commun. ACM. DOI: 10.1145/272682.272711

Michalewicz, Z. and G. Nazhiyath, 1995. Genocop III: A co-evolutionary algorithm for numerical optimization problems with nonlinear constraints. Proceedings of the IEEE International Conference on Evolutionary Computation, 29 Nov-1 Dec, IEEE Xplore Press, Perth, WA, pp: 647-651. DOI: 10.1109/ICEC.1995.487460

Michalewicz, Z. and M. Schoenauer, 1996. Evolutionary algorithms for constrained parameter optimization problems. Evoluti. Comput., 4: 1-32. DOI: 10.1162/evco.1996.4.1.1

Michalewicz, Z. and N. Attia, 1994. Evolutionary optimization of constrained problems. Proceedings of the 3rd Annual Conference on Evolutionary Programming, (CEP' 94), pp: 98-108.

Michalewicz, Z., 1992. Genetic Algorithms + Data Structures = Evolution Program. 2nd Edn., Artificial Intelligence, Springer.

Michalewicz, Z., 1995a. Genetic algorithms, numerical optimization and constraints. Proceedings of the 6th International Conference on Genetic Algorithms, (CGA' 95), Morgan Kaufmann, pp: 151-158.

Michalewicz, Z., 1995b. A survey of constraint handling techniques in evolutionary computation methods. Evolu. Programm., 4: 135-155.

Morales, A.K. and C.V. Quezada, 1998. A universal eclectic genetic algorithm for constrained optimization. Proceedings of the 6th European Congress on Intelligent Techniques and Soft Computing, (TSC' 98), pp: 518-522.

Powell, D. and M.M. Skolnick, 1993. Using genetic algorithms in engineering design optimization with non-linear constraints. Proceedings of the 5th International Conference on Genetic Algorithms, (CGA' 93), pp: 424-431.

Ragsdell, K. and D. Phillips, 1976. Optimal design of a class of welded structures using geometric programming. J. Manufactur. Sci. Eng., 98: 1021-1025. DOI: 10.1115/1.3438995

Rao, S.S., 1996. Engineering Optimization: Theory and Practice. 1st Edn., New Age International, New Dehli, ISBN-10: 8122411495, pp: 112.

Sandgren, E., 1988. Nonlinear integer and discrete programming in mechanical design. Proceedings of the ASME Design Technology Conference, (DTC' 90), pp: 95-105.

Sandgren, E., 1990. Nonlinear integer and discrete programming in mechanical design optimization. J. Mechan. Des., 112: 223-229. DOI: 10.1115/1.2912596

Siddall, J., 1972. Analytical design-making in engineering design. Englewood Cliffs, NJ.

Wu, S.J. and P.T. Chow, 1995. Genetic algorithms for nonlinear mixed discrete-integer optimization problems via meta-genetic parameter optimization. Eng. Optimizat., 24: 137-159. DOI: 10.1080/03052159508941187

Yeniay, O., 2005. Penalty function methods for constrained optimization with genetic algorithms. Math. Comput. Applic., 10: 45-56. DOI: 10.3390/mca10010045

Yun, Y., 2005. Study on adaptive hybrid genetic algorithm and its applications to engineering design problems. Waseda University.